\documentclass[nohyperref]{article}

\usepackage{microtype}
\usepackage{graphicx}
\usepackage{subcaption}
\usepackage{booktabs} 

\usepackage{hyperref}

\usepackage[accepted]{icml2022}

\usepackage{amsmath}
\usepackage{amssymb}
\usepackage{mathtools}
\usepackage{amsthm}
\usepackage{multirow}
\usepackage[capitalize,noabbrev]{cleveref}

\usepackage{makecell}
\theoremstyle{plain}

\theoremstyle{definition}

\theoremstyle{remark}

\usepackage[textsize=tiny]{todonotes}
\DeclareMathOperator*{\NCE}{NCE}

\DeclareMathOperator*{\argmin}{argmin}
\newcommand{\PO}{decision-focused learning}

\DeclareMathOperator*{\LOPSc}{L_{\mbox{OP}^S_c}}
\icmltitlerunning{Decision-Focused Learning: Through the Lens of Learning to Rank}

\begin{document}

\twocolumn[
\icmltitle{Decision-Focused Learning: Through the Lens of Learning to Rank}



\icmlsetsymbol{equal}{*}

\begin{icmlauthorlist}
\icmlauthor{Jayanta Mandi}{vub}
\icmlauthor{V\'ictor Bucarey}{chl}
\icmlauthor{Maxime Mulamba}{vub}
\icmlauthor{Tias Guns}{kul}
\end{icmlauthorlist}

\icmlaffiliation{vub}{Data Analytics Laboratory, Vrije Universiteit Brussel, Belgium}
\icmlaffiliation{chl}{Institute of Engineering Sciences, Universidad de O'Higgins, Rancagua, Chile}
\icmlaffiliation{kul}{Dept. Computer Science, KU Leuven, Belgium}

\icmlcorrespondingauthor{Jayanta Mandi }{jayanta.mandi@vub.be}
\icmlcorrespondingauthor{V\'ictor Bucarey}{victor.bucarey@uoh.cl}
\icmlcorrespondingauthor{Tias Guns}{tias.guns@kuleuven.be}

\icmlkeywords{Learning to rank, Predict and Optimize}

\vskip 0.3in
]



\printAffiliationsAndNotice{}  

\begin{abstract}
In the last years, \PO , also known as predict-and-optimize, 
has received increasing attention. In this setting, the predictions of a machine learning model are used as estimated cost coefficients in the objective function of a discrete combinatorial optimization problem for decision making.
 Decision-focused learning proposes to train the ML models, often neural network models, by directly optimizing the quality of decisions made by the optimization solvers.
 Based on a recent work that proposed a \emph{noise contrastive estimation loss over} a subset of the solution space, we observe that \PO \ can more generally be seen as a learning-to-rank problem, where the goal is to learn an objective function that ranks the feasible points correctly. This observation is independent of the optimization method used and of the form of the objective function. 
 We develop pointwise, pairwise and listwise ranking loss functions, which can be differentiated in closed form given a subset of solutions. 
 We empirically investigate the quality of our generic methods compared to existing \PO \ approaches 
 with competitive results. Furthermore, controlling the subset of solutions allows controlling the runtime considerably, with limited effect on regret.
\end{abstract}

\section{Introduction}
Many real-world decision making problems rely on the combination of machine learning (ML) and combinatorial optimization (CO). In those applications, CO problems are solved to arrive at a decision by maximizing or minimizing an objective function. 
However, often some parameters of the optimization problem, such as costs, prices and profits, are not known but can be estimated from other feature attributes based on historical data. 

A two-stage predict-\emph{then}-optimize approach is widely used by practitioners in both the industry and the public sector, where first an ML model is trained to make point estimates of the uncertain parameters and then the optimization problem is solved using the predictions. 
However, this treats the parameter errors as independent and does not take the \emph{interplay of the parameter errors and their effect} on the combinatorial optimisation problem into account.

Recently \emph{\PO} \citep{aaai/WilderDT19, PogancicPMMR20} (also known as predict-and-optimize~\citep{elmachtoub2021smart, mandi2020smart}) has received increasing attention. In this setting, during training, the ML model is trained with a loss function that first solves the downstream CO problem to observe the joint error. 
On the technical level, the challenge of integrating combinatorial optimization into the training loop of ML is the non-differentiability of combinatorial problems.
To address this, \citet{mulamba2020discrete} recently proposed an approach motivated by noise-contrastive estimation \citep{gutmann2010noise}, where they introduce a new family of surrogate loss functions considering non-optimal feasible solutions as \emph{negative examples}.
In their work, they propose a surrogate loss function, 
which maximizes the divergence between the objective values of the true optimal and the negative examples. 

We argue that \PO \ can be more generally viewed through the lens of learning to rank (LTR) approach \citep{burges2005learning}. In a LTR problem, for each query there is a list of items and their feature variables given.
The training objective is to learn a ranking function, which will rank the (top) items correctly for each query. 
 
In the context of CO problems, a partial ordering of the feasible solutions is induced by the objective function.
Learning the orders of the feasible solution with respect to the objective function hence achieves the goal of \PO.
When compared to the LTR literature, there are two important differences: 1) part of the structure of the objective function is already given, only some of its parameters must be estimated; and 2) the set of `items' are all feasible solutions, which are the same for all CO problem instances (queries), but it is intractable to enumerate all.
While backpropagating, 1) needs not be an issue under the assumption that just the objective function itself is differentiable, which is the case for standard continuous functions including linear functions. With respect to 2), we can subsample the feasible solutions as is done by \citet{mulamba2020discrete}.

The main contributions of this work are the following: First, we formulate the \PO\  as a ranking problem.  Secondly, we introduce and study several well-known learning to rank loss functions. In particular, we formulate pointwise, pairwise and listwise ranking loss functions for \PO \ problems.   
Thirdly, we show that the pairwise ranking loss is a generalization of the loss functions proposed by \citet{mulamba2020discrete}. 
Lastly, we show that in case of a linear objective function, the pointwise and pairwise-difference loss functions can be interpreted as trading off the mean square error (MSE) and regret of predictions.

\section{Related Literature}\label{sect:rl}
\paragraph{Decision-focused learning.} 
Decision-focused learning involves optimization problems,
where the optimization parameters are defined partially. 
One of the major developments in this topic is the \emph{differentiable optimization layer} \citep{amos2017optnet}, which analytically computes the gradients by differentiating the KKT optimality conditions of a quadratic program. The  gradient  calculation  of \citet{amos2017optnet} does not translate to linear programming (LP) problems due to singularities of the objective function in this case. \citet{aaai/WilderDT19} address this by introducing a small quadratic regularizer in the objective function while \citet{mandi2020interior} introduce a log-barrier regularization term and compute the gradients from the homogeneous self dual embedding of the LP. \citet{aaaiFerberWDT20} studies mixed integer LPs and reduce it to LP by adding cutting plane to the root LP node, applying the framework of  \citet{aaai/WilderDT19} afterwards. All these methods are specific to the structure of the CO problem used.

Other approaches are independent of the CO problem and solver used. The \emph{smart predict-and-optimize} approach of \citet{elmachtoub2021smart} studies optimization problems with linear objectives and where the cost vector has to be predicted. They introduce a convex surrogate upper bound of the loss and propose an easy to use subgradient. \citet{PogancicPMMR20} also studies problems with linear objectives. They compute the gradient by perturbing the predicted cost vector. \citet{NEURIPS2021_IMLE} extend this work by adding noise-perturbations for perturbation-based implicit differentiation. For an overview of \PO, we refer the readers to \citet{kotary2021end}.

All these approaches involve solving the optimization problem for every instance during training, which comes at a very high computational cost. To address this problem, \citet{mulamba2020discrete} propose contrastive loss functions with respect to a pool of feasible solutions of the optimization problem. It cleanly separates the solving (adding solutions to the pool), from the loss function. This allows the loss function to be differentiated directly. We build on this idea but through the lens of learning to rank.

\paragraph{Learning to rank.} Learning to rank problems have been studied thoroughly over the years, especially within the context of information retrieval. Readers can refer to \citet{liu2011learning} for a detailed analysis.
Most of the LTR approaches assign real-valued scores to the items, then define surrogate loss functions on the scores.
In pointwise LTR models \citep{li2007mcrank}, the labels are the rank (or true score, if known) of the items. These models fail to consider
any inter dependencies across the item rankings. 
Pairwise ranking approaches \citep{burges2005learning} aim to learn the relative ordering of pairs of items. 
Finally, listwise approaches \citep{cao2007learning} define loss functions with respect to the scores of the whole ranked lists.
The LTR framework has been applied to various contexts such as recommender systems \citep{karatzoglou2013learning} and software debugging \citep{xuan2014learning}, among others.

\citet{emircpaior19} have used LTR in \PO \ framework, but by considering the parameters of a linear objective function as the items to rank, e.g. the value of items of a knapsack problem, instead of the feasible solutions as we do.
\citet{kotary2021fairLTR} propose a fair LTR methodology, which impose fairness constraints using constraints in the combinatorial optimization problem and they use \PO \ to integrate the optimization program with an ML model.
In contrast, we use LTR models and losses for \PO.

\section{Problem Statement and Motivation}
In this section, we introduce the \PO \ setup. We consider a discrete combinatorial optimization problem
\begin{equation}
    v^\star(c) \in \argmin_{v \in V} f(v, c) 
    \label{eq:COP}
\end{equation}
\noindent where $V \subseteq
\mathbb{Z}^K$ is the set of {\em feasible} integer solutions, typically specified implicitly through constraints, and $f:V\times \mathcal{C} \rightarrow \mathbb{R}$ is the real valued objective function, where  $\mathcal{C} \subset \mathbb{R}^K$ is the domain of the vector of parameters $c$.
We denote an optimal solution of \eqref{eq:COP} by $v^\star(c)$.
The value of $c$ is unknown but we have access to correlated features $x$ and a historic dataset $D = \{(x_i, c_i)\}_{i=1}^N$. 
The goal is to predict the apriori unknown coefficient vector $c_i$ in a supervised machine learning setup.
To do so, we train a model 
noted by $m(\omega, x_i)$ to make a prediction of vector $c_i$, where $\omega$ are the model parameters.  Let us denote the predicted value as $\hat{c}_i=m(\omega, x_i)$. 

In a traditional supervised machine learning setup, we train by minimizing the difference between $c$ and $\hat{c}$.
For instance, in a regression problem, we minimize the multi-output mean square error,
\begin{equation}
    \mathbf{mse}(c,\hat{c}) = \sum_{k=1}^K (c_k -\hat{c}_k)^2 = \sum_{k=1}^K \epsilon_k^2 \label{eq:mse_def}
\end{equation}
\noindent where $\epsilon_k = (c_k -\hat{c}_k)$, over a set of $N$ instances $\{(c, \hat{c})\}$.

In contrast, in \PO \ $\hat{c}$ is an intermediate result. The final output is $v^\star(\hat{c})$, the solution of the discrete combinatorial optimization problem. 
The final goal is to minimize the impact of the solution $v^\star(\hat{c})$ in the downstream optimization problem whenever the real cost vector $c$ is realized. 
In order to measure how good $\hat{c}$ is, we compute the \emph{regret} of the combinatorial optimisation. The \emph{regret} is defined as the difference between the value of the optimal objective value $f(v^\star(c),c)$ and the value of the objective function with ground truth $c$ and predicted solution $v^\star(\hat{c})$. Formally, we define regret, without any assumptions on the structure of $f$, as
\begin{equation}
\mbox{Regret}(\hat{c},c) = f(v^\star(\hat{c}), c) -  f(v^\star(c), c). \label{eq:regret}
\end{equation}

Ideally, we aim to learn parameters $\omega$
that determine $\hat{c}=m(\omega, x)$ such that it minimizes the \emph{regret}. 
To do so in a neural network setup, the gradient of the regret has to be backpropagated, which requires computing the exact derivative of the regret \eqref{eq:regret}. This task is problematic as $V$ is discrete and the regret function is non-continuous and involves differentiating over the argmin in $v^\star(\hat{c})$. 


As mentioned in section \ref{sect:rl}, several recent works have studied this problem.
However, in all those cases, in order to compute the gradient,
the optimization problem \eqref{eq:COP} must be solved repeatedly for each instance during training. 
Hence, scalability is a major challenge in these approaches. 
\subsection{Contrastive Surrogate Loss}
To improve the scalability of \PO \ \citet{mulamba2020discrete} proposed an alternative class of loss functions.
They first define the following probability distribution of $v \in V$ being the optimal one.
\begin{equation}
    p(v|c) = \begin{cases}
    \frac{\exp \left( - f \left( v,c \right)\right)}{\sum_{v^{\prime} \in V } \exp \left( - f \left( v^{\prime},c \right)\right) } & \text{if}\ v \in V \\
    0, & \text{if}\ v \notin V
    \end{cases}
    \label{eq:exponential}
\end{equation}
It is easy to verify that, if $v$ is one minimizer of Eq.~\ref{eq:COP}, then it will maximize Eq.~\ref{eq:exponential}.
To define the loss functions they consider to have access to $S$, a \emph{subset} of $V$ and treat $\left( S \setminus \{v^\star(c)\} \right)$ as negative examples. With this setup, they define the following noise contrastive estimation (NCE) loss
\begin{align} 
		L_{\NCE}(\hat{c},c) = \frac{1}{|S|} \sum_{v^s \in S} \Big(f\big(v^\star(c),\hat{c} \big) -  f\big(v^s, \hat{c} \big) \Big). 
		\label{eq:nce}
\end{align}
The novelty of this approach is that Eq.~\eqref{eq:nce} is differentiable and that the differentiation does not involve solving the optimization problem in \eqref{eq:COP}. Further, if the solutions in $S$ are optimal solutions of arbitrary cost vectors, this approach is equivalent to training in a region of the convex-hull of $V$.

\section{Rank Based Loss Functions}\label{sect:lossfunctions}
In this section we develop and motivate a family of surrogate loss functions for \PO \ problems. 
We remark that, for a given $c$, we can create a partial ordering of all $v \in V$ by ordering them with respect to the objective value.
Let us denote by $p_1, \ldots ,p_{|V|}$ the indices of $V$ so that $f(v^{p_1},c ) \leq f(v^{p_2},c ) \leq \ldots \leq f(v^{p_{|V|}},c )$. 
The key idea of our approach is to generate prediction $\hat{c}$ so that $f(v^{p_1},\hat{c}) \leq f(v^{p_2},\hat{c} ) \leq \ldots \leq f(v^{p_{|V|}},\hat{c} )$. 
Notice that, if the ranks of $v \in V$ with respect to $\hat{c}$ are identical to that of $c$, the regret is zero. Furthermore, it is sufficient that $\forall p_s: f(v^{p_1},\hat{c}) \leq f(v^{p_s},\hat{c} )$ for the regret to be zero. 
In general, our surrogate task is to learn the ranking of each $v \in V$ with respect to $c$. This motivates us to use the LTR framework to develop a new class of surrogate loss functions for  \PO \ problems.

Drawing a parallel to the LTR literature, we consider each $x$ a query, and the feasible solutions $v \in V$ as the set of items to be ordered. 
Our formulation differs from traditional LTR framework because in LTR problems each item has its own set of feature variables, whereas in our formulation each $v$ does not have feature variables. We hence only have query-features ($x$ itself) to predict $c$. 
In the LTR framework, a ranking function is used to assign scores to each item. 
In \PO setup, we will consider the objective function $f(v,c)$ as the scoringh function.  

To rank all $v \in V$ is an intractable task for most combinatorial problems. To overcome this in practice, we will consider a set of feasible solution $S \subset V$.
The proposed implementation is described in pseudocode form in Algorithm~\ref{Alg}. For each $c$, we compute a loss function $L(\hat{c},c)$ considering the surrogate ranking task.
The implementation details are described later. Next, we will introduce the family of rank based loss functions.

\begin{algorithm}[tb]
\caption{Gradient-descent implementation of  \PO \ problems with Ranking Loss}
\textbf{Input}: D$ \equiv \{(x_i,c_i)\}_{i=1}^n$
\begin{algorithmic}[1] 
\STATE Initialize $\omega$
\STATE Initialize $S= \{v^\star(c_i) | (x_i, c_i) \in D \}$  \label{alg:pool_initialize} \\
\textbf{Training}
\FOR{each epochs}
\FOR{each $(x_i,c_i)$}
\STATE $\hat{c}_i = m (\omega, x_i)$\\
\IF{random$() < p_{solve}$} \label{alg:ln:growth}
    \STATE $v^\star (\hat{c}_i) \leftarrow $call solver to solve Eq. \eqref{eq:COP},  
    \STATE $S \leftarrow S \cup \{v^\star (\hat{c}_i) \}$ \label{alg:pool_growth}
\ENDIF
\STATE $\mathcal{L} += L(\hat{c}_i, c_i;S)$
\ENDFOR
\STATE $\omega \leftarrow \omega - \alpha \frac{\partial \mathcal{L}}{\partial \omega}$ \# backward pass
\ENDFOR
\end{algorithmic}
\label{Alg}
\end{algorithm}


\subsection{Pointwise Ranking}
In pointwise ranking loss formulation, the loss function is defined for each item (feasible solution) independently.
In our pointwise rank loss formulation, we propose to regress the predicted objective function value on the actual objective function value. Thus, we propose the \textbf{pointwise loss} function 
\begin{align}
    L_P(\hat{c},c; S) = \frac{1}{|S|} \sum_{ v \in S} \left( f(v, \hat{c}) - f(v, c) \right)^2
    \label{eq:pointwise}
\end{align}
that measures the errors in the evaluation of the predicted cost function at each point in $S$. 

\paragraph{Intuitive idea of the pointwise loss function.}
When the objective function is linear i.e. $f(v,c) = c^\intercal v$, the pointwise loss~\eqref{eq:pointwise} becomes
\begin{align}
    L^p(\hat{c},c; S) =\frac{1}{|S|} \sum_{v \in S} \left( c^\intercal v - \hat{c}^\intercal v  \right)^2
\end{align}
The derivative of this loss is
\begin{align}
    \frac{\partial L^p(\hat{c},c; S)}{\partial \hat{c} } =-\frac{2}{|S|} \sum_{v \in S} \left( c^\intercal v  - \hat{c}^\intercal v  \right)v
    \label{eq:pointwise_grad}
\end{align}
When considering a combinatorial problem where $v$ is a binary vector, we can see from  Eq.~\eqref{eq:pointwise_grad} that
coordinates where $v$ has zeros, will not contribute to the gradient. 
Moreover, the first component
$\left( c^\intercal v - \hat{c}^\intercal v  \right)$ is the difference between the actual and the predicted objective value on $v$. The gradient contribution of $v$ is weighed by this difference. 

\paragraph{Pointwise loss as a regularizer.}
For the special case of a linear objective function 
$f(v,c) =$ $\sum_{j=1}^K  c_j v_j$, the pointwise task loss can be rewritten as:
\begin{align}
    L_P(\hat{c},c; S)
    =& \sum_{i=1}^K (\hat{c}_i-c_i)^2 \gamma_i +\sum_{i\neq j} (\hat{c}_i-c_i)(\hat{c}_j-c_j) \gamma_{ij} \nonumber \\
    =&\sum_{i=1}^K \epsilon_i^2 \gamma_i +\sum_{i\neq j} \epsilon_i\epsilon_j \gamma_{ij} \label{eq:point_wise_linear}
\end{align}

\noindent where $\gamma_i = \frac{1}{|S|}\sum_{v\in S} v_i^2$ and $\gamma_{ij} = \frac{1}{|S|}\sum_{v\in S} v_i v_j$ and $\epsilon_i$ defined as in \eqref{eq:mse_def}. In this case, 
it is a weighted version of the $\mathbf{mse}$ function plus a term that measures the pairwise relationship of errors of the coordinates. 

Moreover, when $V\subseteq \{0,1\}^K$, the weight $\gamma_i$ is the frequency of how many times coordinate $v_i$ takes value 1 (since $v_i^2 = v_i$). Coefficients $\gamma_{ij}$ measure the frequency where $v_i$ and $v_j$ take value 1 at the same time. In particular, if $S=V$
\begin{align}
    L^p(\hat{c},c; V) = \frac{1}{2} \Big( \mathbf{mse}(\hat{c},c) + 
    \sum_{i=1}^K\sum_{j=1: j< i}^K \epsilon_i \epsilon_j
    \Big) \nonumber
\end{align}
Hence, the pointwise loss function can be seen as the $\mathbf{mse}$ plus a regularizer which penalizes the crossed-errors between coordinates of the vector $c$.

\subsection{Pairwise Ranking}\label{sect:pairwise}
{\em Example.} To demonstrate the shortcoming of the pointwise loss and the motivation behind the pairwise loss formulation, we construct a simple example. Let us consider the feasible space is the 2 dimensional 0-1 space, containing the four feasible points $[0,0],[0,1],[1,0],[1,1]$. 
Now, let $c$ be $[2,-5]$. The objective values at these four points are $0,-5,2$ and $-3$, hence $v^\star(c) = [0,1]$. Now let $\hat{c}_1$ be $[-1,1]$ and $\hat{c}_2$ be $[5,-11]$.
For  $\hat{c}_1$ and $\hat{c}_2$, the objective values of these four points are $0,1,-1,0$ and $0,-11,5,-6$ respectively.
The square error of both $\hat{c}_1$ and $\hat{c}_2$ is $45 =(3^2+6^2)$. The pointwise loss between $c$ and $\hat{c}_1$ is $54=(0+(-5-1)^2+(2+1)^2+(-3-0)^2)$ and $c$ and $\hat{c}_2$ is $54= (0+(-5+11)^2+(2-5)^2+(-3+6)^2)$. But $v^\star(\hat{c}_1) = [1,0]$ and
$v^\star(\hat{c}_2) = [0,1] = v^\star(c)$.
Even though the regret of $\hat{c}_2$ is $0$, its pointwise loss is the same as $\hat{c}_1$, whose regret is positive. With this motivation behind, we will construct the pairwise ranking loss function.

Let $(p,q)$ be an ordered pair in $S$ with respect to $c$. We define $\mbox{OP}_c^S$ as the set of all the ordered pairs such that the first coordinate dominates the second one in the order induced by $c$, or equivalently:
\begin{equation}
    (p,q) \in \mbox{OP}_c^S \implies f(v^p,c ) < f(v^q,c) 
    \label{eq:ordered pair}
\end{equation}
For an ordered pair $(p,q)$, we expect the predicted $\hat{c}$ would also respect the ordering i.e. $f(v^p,\hat{c} ) < f(v^q,\hat{c} ) $. In other words, $f(v^p,\hat{c} ) - f(v^q,\hat{c} )$ is the quantity we aim to minimize.
In the standard pairwise \emph{margin} ranking loss formulation \citep{Joachims02}, the loss is zero, if $f( v^p,\hat{c})$ is smaller than $f( v^q,\hat{c})$ by at least a margin of $\nu >0$. 
With this notion of margin, we introduce the following generic \textbf{pairwise loss}
\begin{align}
     & \LOPSc(\hat{c},c)  = \nonumber \\ & \frac{1}{|\mbox{OP}_{c}^S|}  \sum_{(p,q) \in \mbox{O.P.}_{c}^S}  \max \left(0,  \nu+ f( v^p,\hat{c}) - f( v^q,\hat{c}) \right) 
    \label{eq:pairwise}
\end{align}

\paragraph{Ordered pairs generation.} 
Eq.~\eqref{eq:pairwise} is built upon the concept of a pairwise ranking loss \citep{burges2005learning}. 
Note that, to implement this approach we have to identify ordered pairs from the negative examples.  Clearly generating all possible pairs comes with a complexity of $\mathcal{O}(|S|^2)$. 
To overcome this implementation challenge, we consider \emph{best-versus-rest} scheme, i.e.
 $\mbox{OP}_{c}^S = \{ ( v^{p_1}, v^{p_2} ), ( v^{p_1}, v^{p3} ), \ldots ,( v^{p_1}, v^{p_S} )\}$, where $v^{p_1}= v^\star(c)$.
With this \emph{best-versus-rest} pair generation scheme and $\gamma$ being $0$, Eq.~\eqref{eq:pairwise} becomes
\begin{equation}
  \frac{1}{|S|} \sum_{v^q \in S}  \max \left(0,( f( v^\star(c),\hat{c}) - f( v^q,\hat{c}) ) \right)
    \label{eq:pairwiseranking_loss}
\end{equation}
We point out that \eqref{eq:pairwiseranking_loss} takes the form of \eqref{eq:nce}, without the \emph{relu} operator. This shows that the NCE approach of \citet{mulamba2020discrete}, although derived differently, can be seen as a special case of our \emph{best-versus-rest} pairwise approach.
    
{\em Example continued.}
For computing the pairwise loss, first, we have to generate the ordered pairs. The ordering of the feasible points with respect to $c$ is $[0,1],[1,1],[0,0],[1,0]$. 
Following the best-vs-rest scheme, the pairs are $\{ ([0,1],[1,1]), ([0,1],[0,0]),([0,1],[1,0]) \}$.
For $\hat{c}_1$ the value of the pairwise loss is $1+1+2 =4$.
For $\hat{c}_2$ the value of pairwise loss is $0+0+0=0$, which is reasonable as its regret is $0$.
\subsection{Regress on Pairwise Difference}
Another way to generate a ranking of solutions in $S$ is to generate predicted cost vectors, that produce the same {\em difference} in objective value.
To do so, we propose a loss function over the difference over pairs of solutions their cost vectors $\hat{c}_i$ and $c_i$. This \textbf{pairwise difference loss} function can be formulated as:
\begin{align}
     L_{pd}(\hat{c},c; S) =  \frac{1}{|S|}  &\sum_{(p,q) \in \mbox{OP}_{c}^S}  \bigg( \left( f( v^p,\hat{c}) -   f( v^q,\hat{c}) \right) - \nonumber  \\
     &\left( f( v^p,c) - f( v^q,c) \right) \bigg)^2
    \label{eq:pairwise_diff}
\end{align}
For the sake of clarity, we ommitted $\mbox{O.P.}_{c}^S$ from the definition of $L_pd$. 

It is easy to check that in the case of a linear objective:
\begin{align}
    L_{pd}(\hat{c},c; S) = &\sum_{i=1}^K \epsilon_i^2 \overline{\gamma}_i +\sum_{i\neq j} \epsilon_i\epsilon_j \overline{\gamma}_{ij} \label{eq:pair_wise_difference}
\end{align}
\noindent where $\overline{\gamma}_i =\frac{1}{|S|} \sum_{(p,q)}(v^p_i-v^q_i)^2 $, which is proportional to how many times $v^p_i$ and $v^q_i$ are different,
and $\overline{\gamma}_{ij} = \frac{1}{|S|} \sum_{(p,q)}(v^p_i-v^q_i)(v^p_j-v^q_j)$. As in the pointwise loss, the pairwise difference loss can be seen as a weighted $\mathbf{mse}$ loss function with a regularizer term; this time involving the difference between solutions rather then individual coefficients. 


{\em Example continued.}
As before, the set of pairs is $\{ ([0,1],[1,1]), ([0,1],[0,0]),([0,1],[1,0]) \}$. The difference between these three sets of pairs for $c$ are $2,5,7$. The corresponding differences for $\hat{c}_1$ and $\hat{c}_2$ are  $-1,-1,-2$ and $5,11,16$.
So, the corresponding losses for $\hat{c}_1$ and $\hat{c}_2$ are $(2+1)^2+(5+1)^2+(7+2)^2 =94$ and $(2-5)^2+(5-11)^2+(7-16)^2 =94$.

\subsection{Listwise Ranking}
Instead of considering instance pairs, the listnet loss  \citep{cao2007learning} considers the entire ordered list. It is obvious that, if the list contains only two instances, then there are no differences between the two approaches. 
The listnet loss computes the probabilities of each item being the top one and then defines a cross entropy loss of these probabilities. 
Instead of using Eq.~\eqref{eq:exponential} directly, we further introduce a temperature parameter $\tau$~\citep{NEURIPS2021_IMLE}  to define
\begin{align}
    p_{\tau}(v|c) = 
    \frac{\exp \left( - \frac{f \left( v,c \right)}{\tau} \right)}{\sum_{v^{\prime} \in V } \exp \left( \frac{- f \left( v^{\prime},c \right)}{\tau} \right) } & \ \forall v \in V
    \label{eq:temp_exponential}
\end{align}
In this formulation, $\tau$ controls the smoothness of the distribution. As $\tau \to 0$, $ p_{\tau}(v|c)$ takes the form of a categorical distribution with \emph{only} $v^\star(c)$ having positive probability mass. 
As $\tau$ increases $ p_{\tau}(v|c)$ converges to a uniform distribution in $V$.

In order to avoid the computation of the partition function (denominator), we will approximate it as in the ranking functions proposed above, namely to replace $V$ by a subset $S \subseteq V$.
 
Given a probability distribution, listnet loss is the cross entropy loss between the predicted and the target distribution. Following this idea, we define \textbf{listwise loss} as the cross entropy loss between $p_{\tau}(v | c)$ and $p_{\tau}(v | \hat{c})$ for all $v \in S$:
\begin{align}
    L^l_S(\hat{c},c) = -\frac{1}{|S|} \sum_{v \in S} p_{\tau}(v|c) \log p_{\tau}(v|\hat{c})
    \label{eq:listwise}
\end{align}
We remark that we could also consider the KL divergence loss instead of cross entropy loss and define listwise loss as 
\begin{align*}
    \frac{1}{|S|} \sum_{v \in S} p_{\tau}(v|c) \bigg( \log p_{\tau}(v|c) - \log p_{\tau}(v|\hat{c}) \bigg)
\end{align*}
Using this loss formulation, we obtained similar results to Eq.\eqref{eq:listwise} and do not further discuss it.
\begin{figure*}[bt]
    \begin{subfigure}[b]{0.25\linewidth}
        \centering
        \includegraphics[scale=0.2]{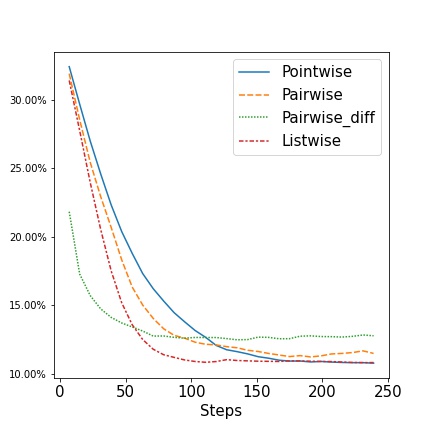}
        \label{}
        \caption{Shortest path (Deg= 2)}
    \end{subfigure}%
    \begin{subfigure}[b]{0.25\linewidth}
        \centering
        \includegraphics[scale=0.2]{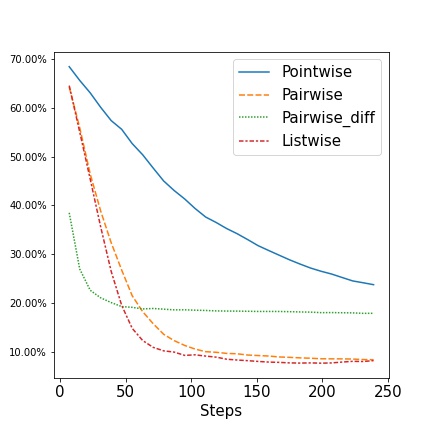}
        \label{}
        \caption{Shortest path (Deg= 4)}
    \end{subfigure}%
    \begin{subfigure}[b]{0.25\linewidth}
        \centering
        \includegraphics[scale=0.2]{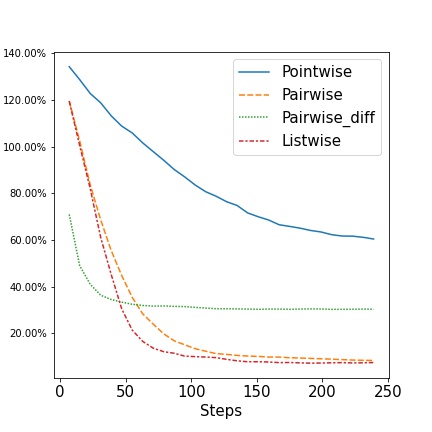}
        \label{}
        \caption{Shortest path (Deg= 6)}
    \end{subfigure}%
    \begin{subfigure}[b]{0.25\linewidth}
        \centering
        \includegraphics[scale=0.2]{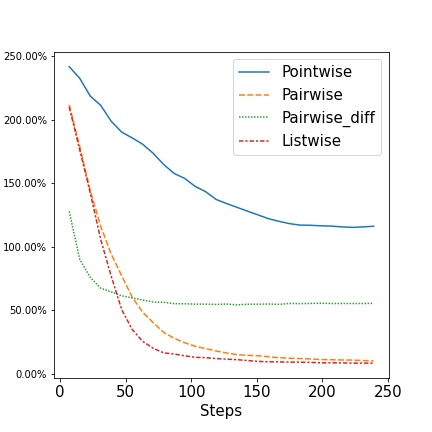}
        \label{}
        \caption{Shortest path (Deg= 8)}
    \end{subfigure}
    \begin{subfigure}[b]{0.25\linewidth}
        \centering
        \includegraphics[scale=0.2]{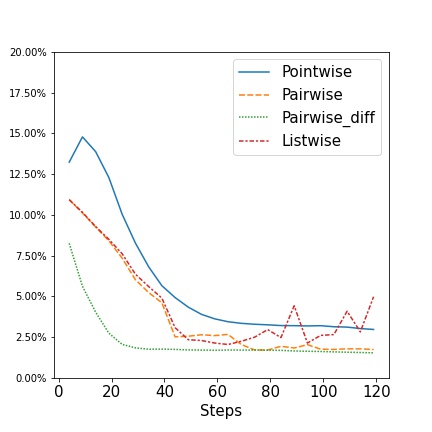}
        \label{}
        \caption{Energy-1}
    \end{subfigure}%
    \begin{subfigure}[b]{0.25\linewidth}
        \centering
        \includegraphics[scale=0.2]{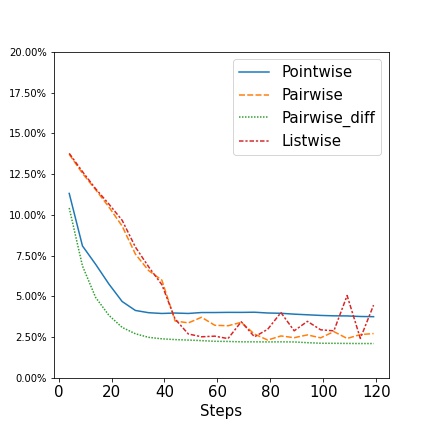}
        \label{}
        \caption{Energy-2}
    \end{subfigure}
    \begin{subfigure}[b]{0.25\linewidth}
        \centering
        \includegraphics[scale=0.2]{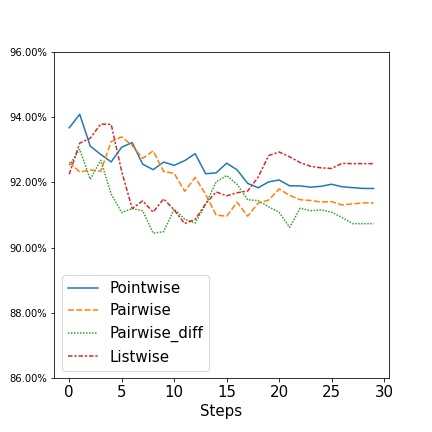}
        \label{}
        \caption{Matching-1}
    \end{subfigure}%
    \begin{subfigure}[b]{0.25\linewidth}
        \centering
        \includegraphics[scale=0.2]{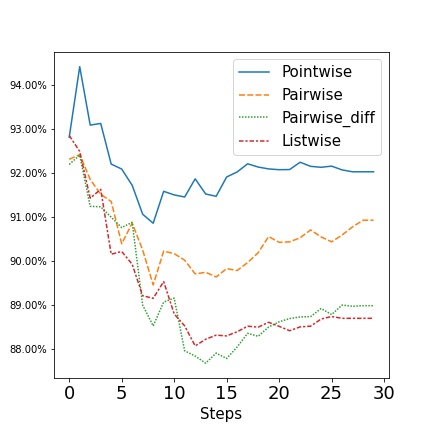}
        \label{}
        \caption{Matching-2}
    \end{subfigure}
    \caption{Learning Curve. We show how the average regrets on validation dataset progress as the models learn. }
    \label{fig:learning_curve}
\end{figure*}
\subsection{Implementation}\label{sect:implementation}
So far we define pointwise \eqref{eq:pointwise} pairwise \eqref{eq:pairwise}, pairwise-difference \eqref{eq:pairwise_diff} and listwise \eqref{eq:listwise} ranking losses. 
As mentioned, the fundamentals of the LTR loss functions is to learn the partial ordering of $v \in S$. 
To construct and expand $ S$, we use the strategy of \citet{mulamba2020discrete}. The generic structure of the proposed algorithm is shown in Algorithm~\ref{Alg},
 which is a modified version of standard gradient descent algorithms for  \PO \ problems. 
They proposed to initialise $S$ with all true optimal solutions. 
In Algorithm~\ref{Alg} $S$ is initialized by caching all the optimal solutions of the training instances in line 2. To expand $S$, 
they defined a hyperparameter $p_{solve}$, which is the probability of calling a optimization oracle to solve an instance during training.  If an instance is solved, its solution is added to $S$. This is demonstrated in line 8 in Algorithm~\ref{Alg}. 
Obviously with $p_{solve} = 100 \% $, Eq.~\eqref{eq:COP} is solved for every $\hat{c}_i$ during training. 

\begin{table*}[ht]
\caption{Comparison of regret on three problem sets. We present average percentage regret of 10 runs.}
\label{tab:Compa}
\begin{small}
\begin{center}
\label{tab:}
\begin{tabular}{@{}cc cc c|c ccc|cc@{}}
\toprule
    & Pointwise &  Pairwise & Pairwise-diff & Listwise & NCE & MAP & SPO & Blackbox & Twostage\\
    \bottomrule
    \multicolumn{10}{c}{Shortest Path Problem}\\
    \toprule
Deg-1 & \textbf{15.45\%} & 16.37\% & 16.00\% & 15.56\% & 23.63\% &19.68\% & 15.51\% & 18.62\% & \textbf{15.45\%}\\
\midrule
Deg-2 & 10.07\% & 10.74\% & 11.60\% & 10.41\% & 30.84\% & 14.15\% & \textbf{10.06\%} & 13.26\% & 10.07\% \\
\midrule
Deg-4 & 19.27\% & 8.06\% & 17.89\% & \textbf{7.71\%} & 65.10\% & 9.44\% & 7.98\% & 9.39\% & 10.44\%\\
\midrule
Deg-6 & 72.18\% & 9.27\% & 32.63\% & \textbf{8.29\%} & 129.42\% & 9.45\% & 16.19\% & 9.69\% & 15.96\%\\
\midrule
Deg-8 &  178.16\% & 12.75\% & 61.80\% & \textbf{12.38\%}  & 242.30\%& 13.68\% & 33.75\% & 14.14\% & 28.50\%\\
    \bottomrule
    \multicolumn{10}{c}{Energy Scheduling Problem}\\
    \toprule
Energy-1 &  2.51\% & 1.65\% & 1.63\% & 1.67\% & 1.69\% & 1.59\% & 1.56\% & \textbf{1.54\%} & 1.91\% \\
\midrule
Energy-2 &  2.83\% & 2.08\% & 2.04\% & 1.96\% & 2.23\% & 2.01\% & \textbf{1.93\%} & \textbf{1.93\%} & 2.46\%\\
    \bottomrule
    \multicolumn{10}{c}{Bipartite Matching Problem}\\
    \bottomrule
Matching-1 & 90.17\% & 88.80\% & \textbf{88.23\%} & 88.49\% & 89.39\% &    89.13\% & 88.70\% & 90.09\% & 90.12\%\\
\midrule
Matching-2 & 89.86\% & 87.24\% & 85.87\% & 86.40\% & 85.31\%  &  \textbf{84.80\%} & 85.79\% &  90.31\% & 89.42\%\\
\bottomrule
\end{tabular}
\end{center}
\end{small}
\end{table*}
The distinctive feature of this proposed approach is that the loss function $L(\hat{c}_i,c_i)$ does not involve any argmin differentiation. So we can use standard automatic differentiation libraries and it can be run on GPU. Yet, it still considers the task loss of the downstream optimization problem. The part that can not be run on GPU is calling an optimization oracle to expand $S$.

\section{Experiments} 
We analyze our approaches on a series of experiments. We consider three discrete combinatorial optimization problems.
First, we briefly describe the experimental setups.
\paragraph{Shortest Path Problems.}
We adopt this synthetic experiment from \citet{elmachtoub2021smart}. It is a shortest path problem in a $5 \times 5$ grid, with the objective of going from the southwest corner of the grid to the northeast corner where the edges can go either north or east. The feature vectors are sampled from a multivariate Gaussian distribution. They generate the cost vector as per the following model
\begin{equation}
    c_{ij} = \bigg(\frac{1}{\sqrt{p}} \big(B x_i  \big) +3 \bigg)^{\text{Deg} }\epsilon_i^j
\end{equation}
where $c_{ij}$ is the $j$-th component of cost vector $c_i$ and $B$ is the latent data generation matrix. Deg specifies the extent of model specification, the higher the value of Deg, the more it deviates from a linear model and the more errors there will be. We experiment with degree of $1,2,4, 6 \text{ and } 8$.
$\epsilon_i^j$ is a multiplicative noise that is sampled from $ \text{U} (0.5,1.5)$. 
For each degree, we have 1000, 250 and 10,000 training, validation and test instances\footnote{The dataset is generated as described in \url{https://github.com/paulgrigas/SmartPredictThenOptimize}}. The underlying model is a neural network without any hidden layer i.e. a regression model.
\paragraph{Energy-cost Aware Scheduling.}
Next we consider the problem of energy-cost aware scheduling, adapted from \emph{CSPLib}~\cite{csplib}, a library of constraint optimization problems.
In energy-cost aware scheduling \cite{csplib:prob059}, a given number of tasks, 
to be scheduled on a certain number of machines without violating the resource capacities of each machine. 
The goal of the optimization problem is to find the scheduling which would minimize the total energy consumption subject to the energy prices, which would be realized in the future. The prediction task is to predict the future energy prices using feature variables 
such as calendar attributes; day-ahead estimates of weather characteristics, forecasted energy-load, wind-energy production and prices.
We take the energy price data from \citet{ifrim2012properties}. 
We study two instances named Energy-1 and Energy-2. The first one involves scheduling 3 tasks on 10 machines, whereas the other instance has 3 tasks and 15 machines.
\begin{figure*}[ht]
    \begin{subfigure}[b]{0.5\linewidth}
        \centering
        \includegraphics[scale=0.35]{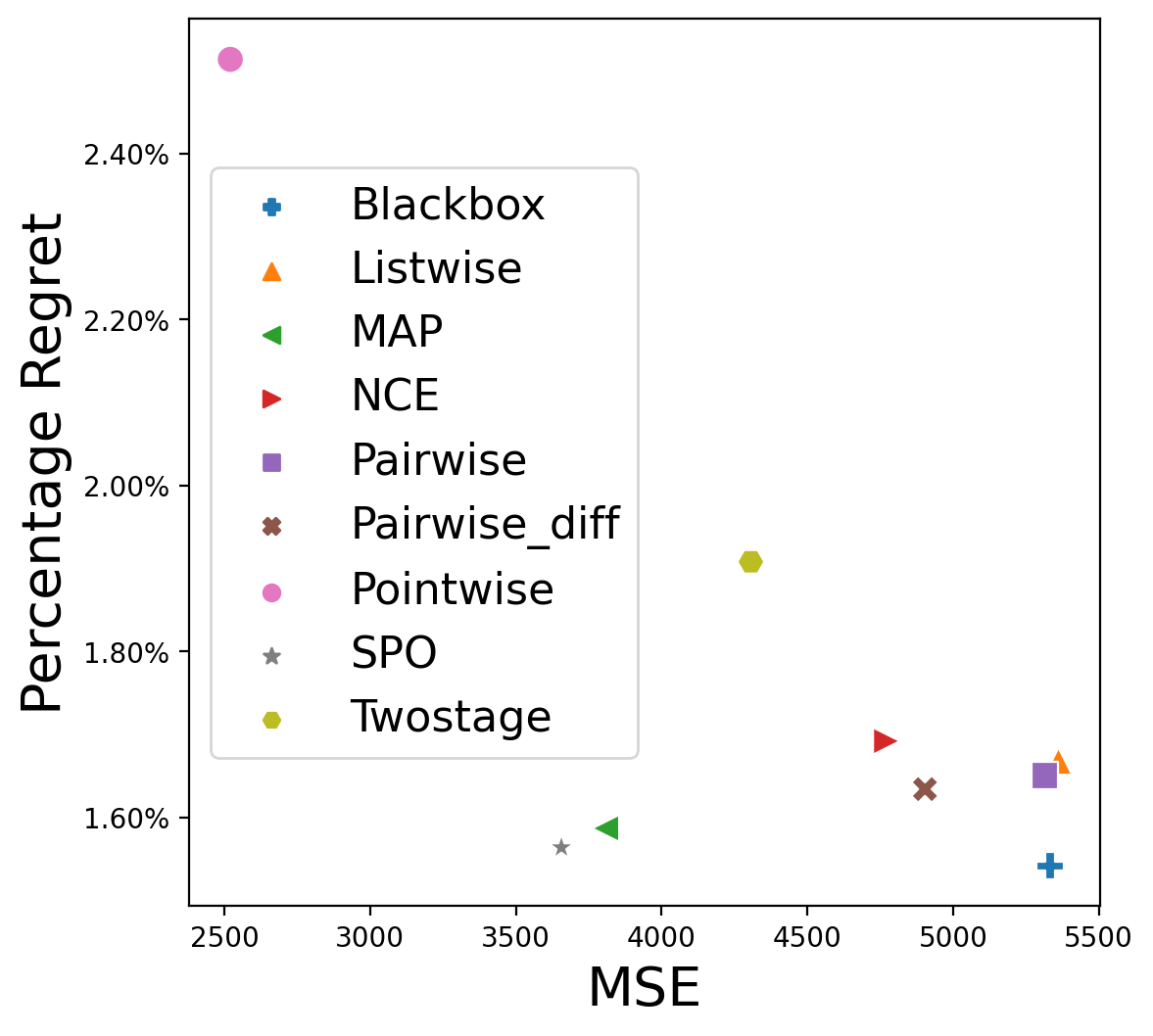}
        \label{}
        \caption{Energy-1}
    \end{subfigure}%
    \begin{subfigure}[b]{0.5\linewidth}
        \centering
        \includegraphics[scale=0.35]{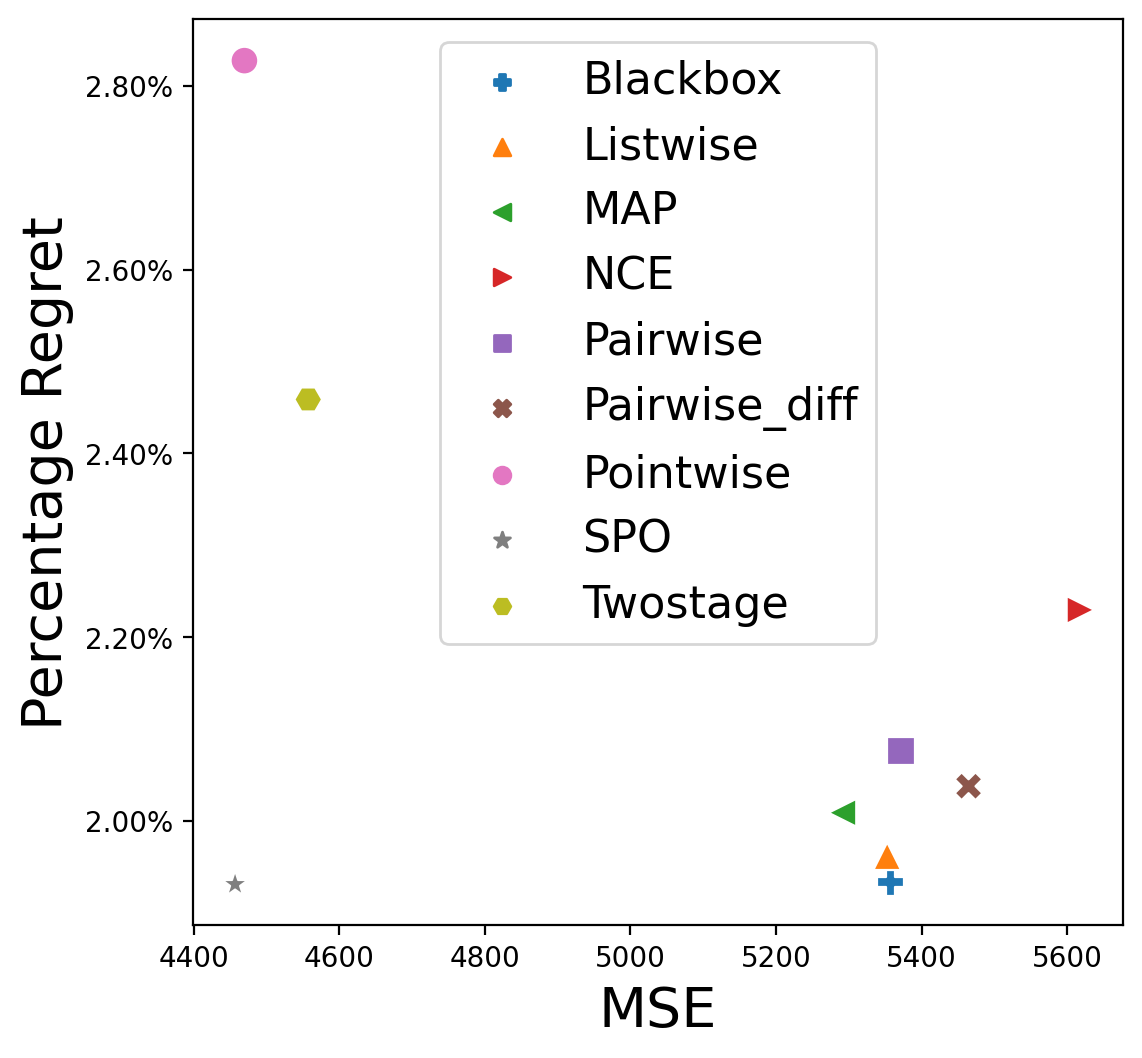}
        \label{}
        \caption{Energy-2}
    \end{subfigure}
        \begin{subfigure}[b]{0.5\linewidth}
        \centering
        \includegraphics[scale=0.35]{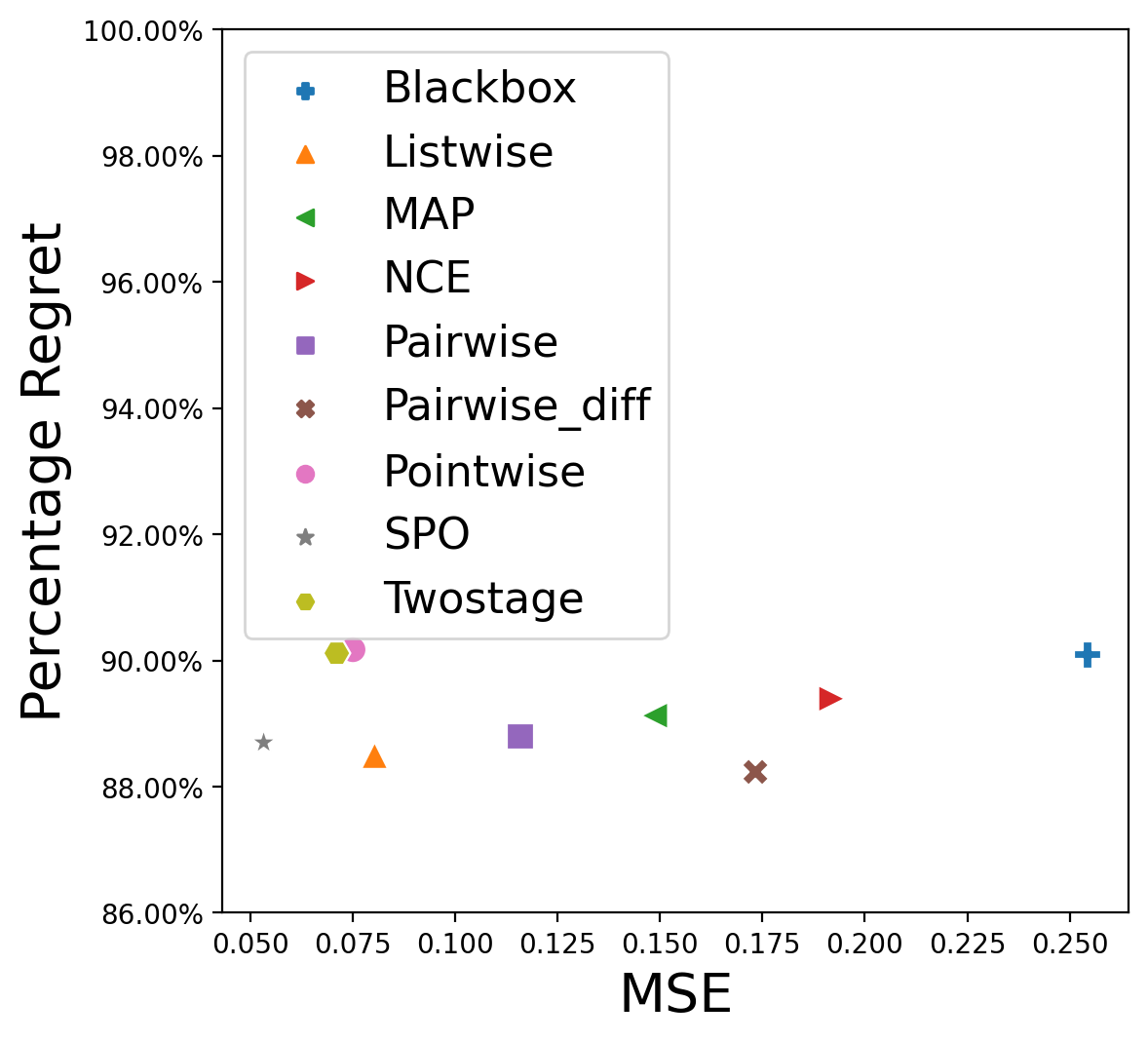}
        \label{}
        \caption{Matching-1}
    \end{subfigure}%
    \begin{subfigure}[b]{0.5\linewidth}
        \centering
        \includegraphics[scale=0.35]{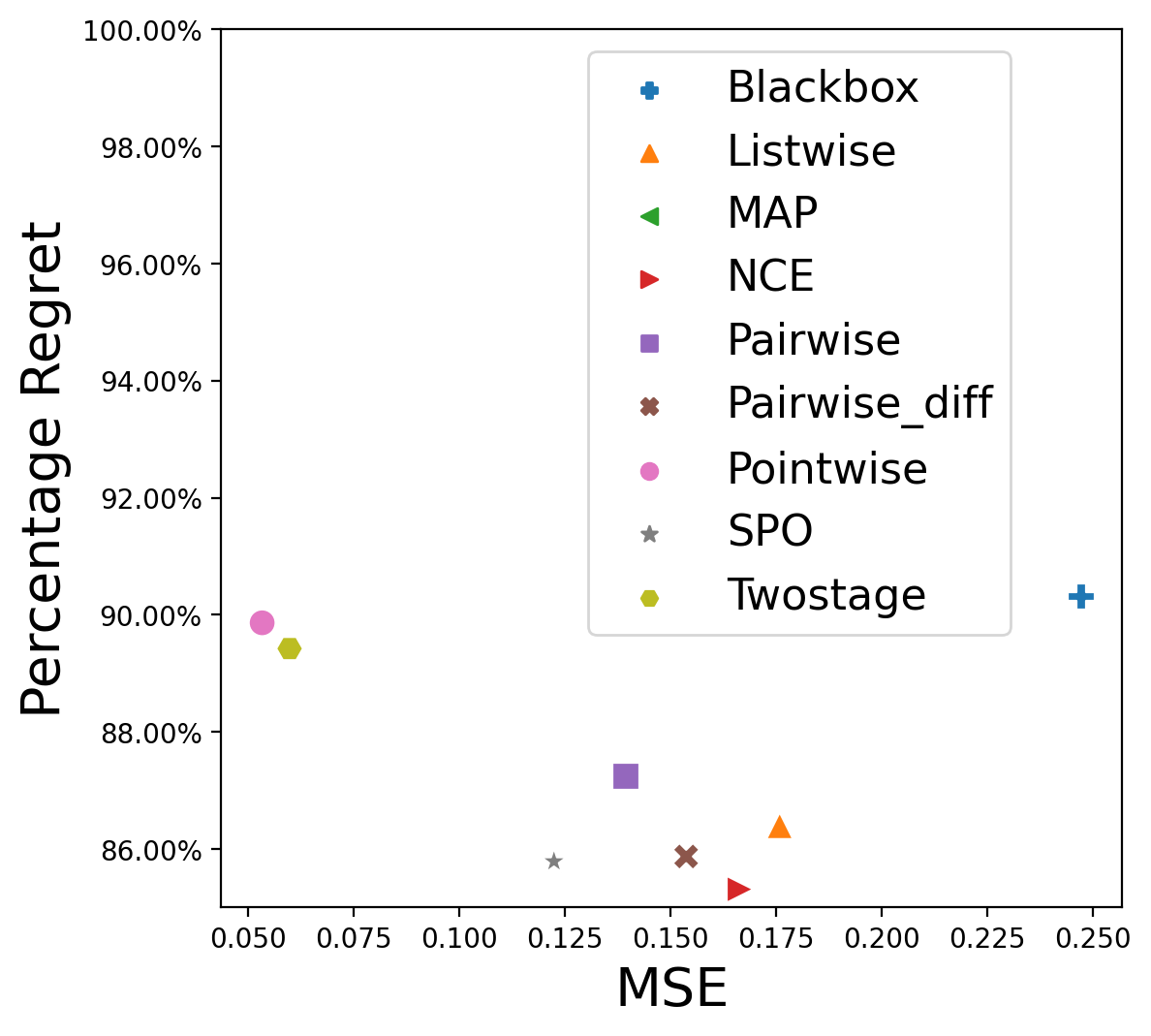}
        \label{}
        \caption{Matching-2}
    \end{subfigure}%
    \caption{MSE vs Regret trade-off}
    \label{fig:msevsregret}
\end{figure*}
\begin{figure*}[ht]
    \begin{subfigure}[b]{0.5\linewidth}
        \centering
        \includegraphics[scale=0.3]{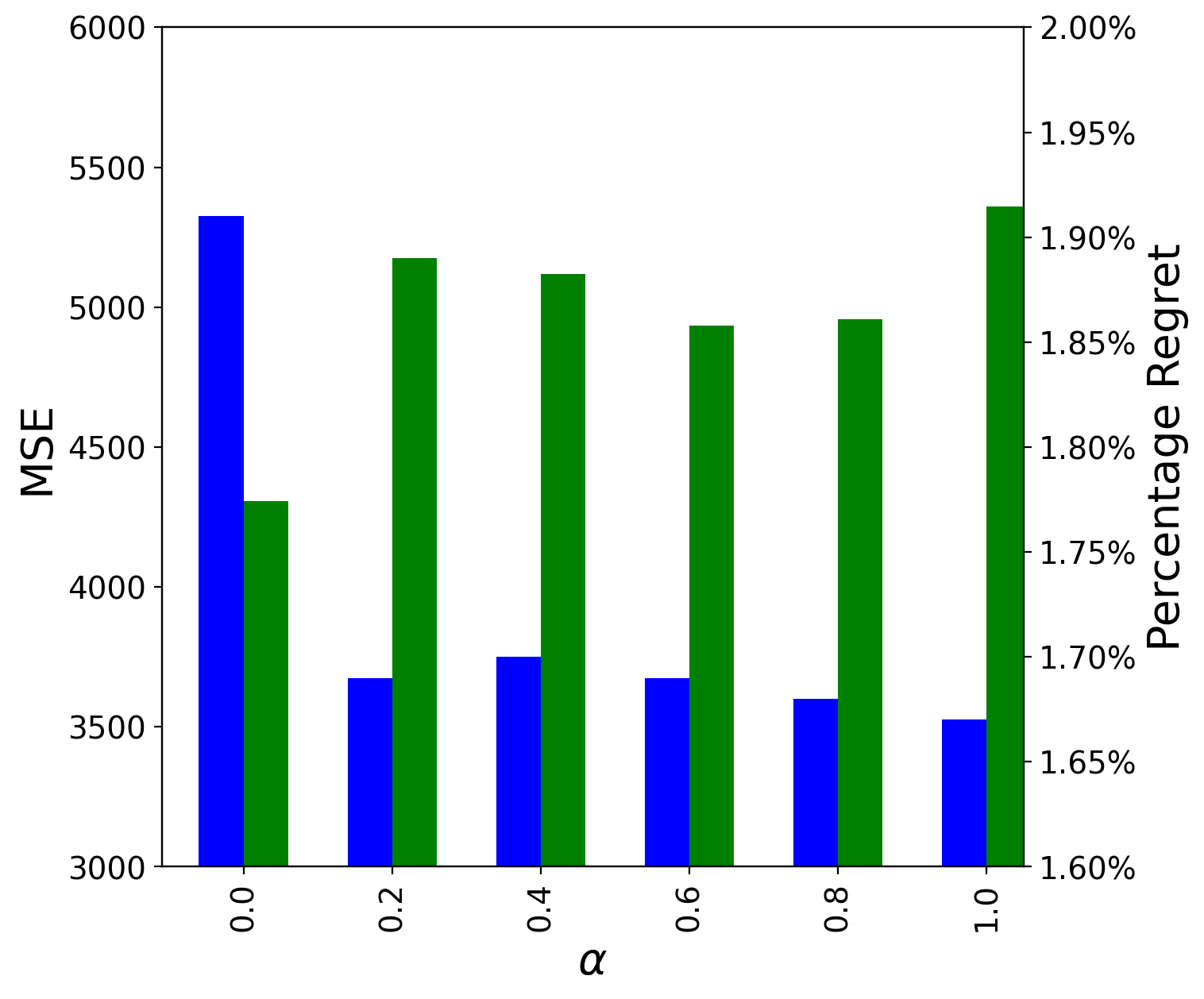}
        \label{}
        \caption{Energy-1}
    \end{subfigure}%
    \begin{subfigure}[b]{0.5\linewidth}
        \centering
        \includegraphics[scale=0.3]{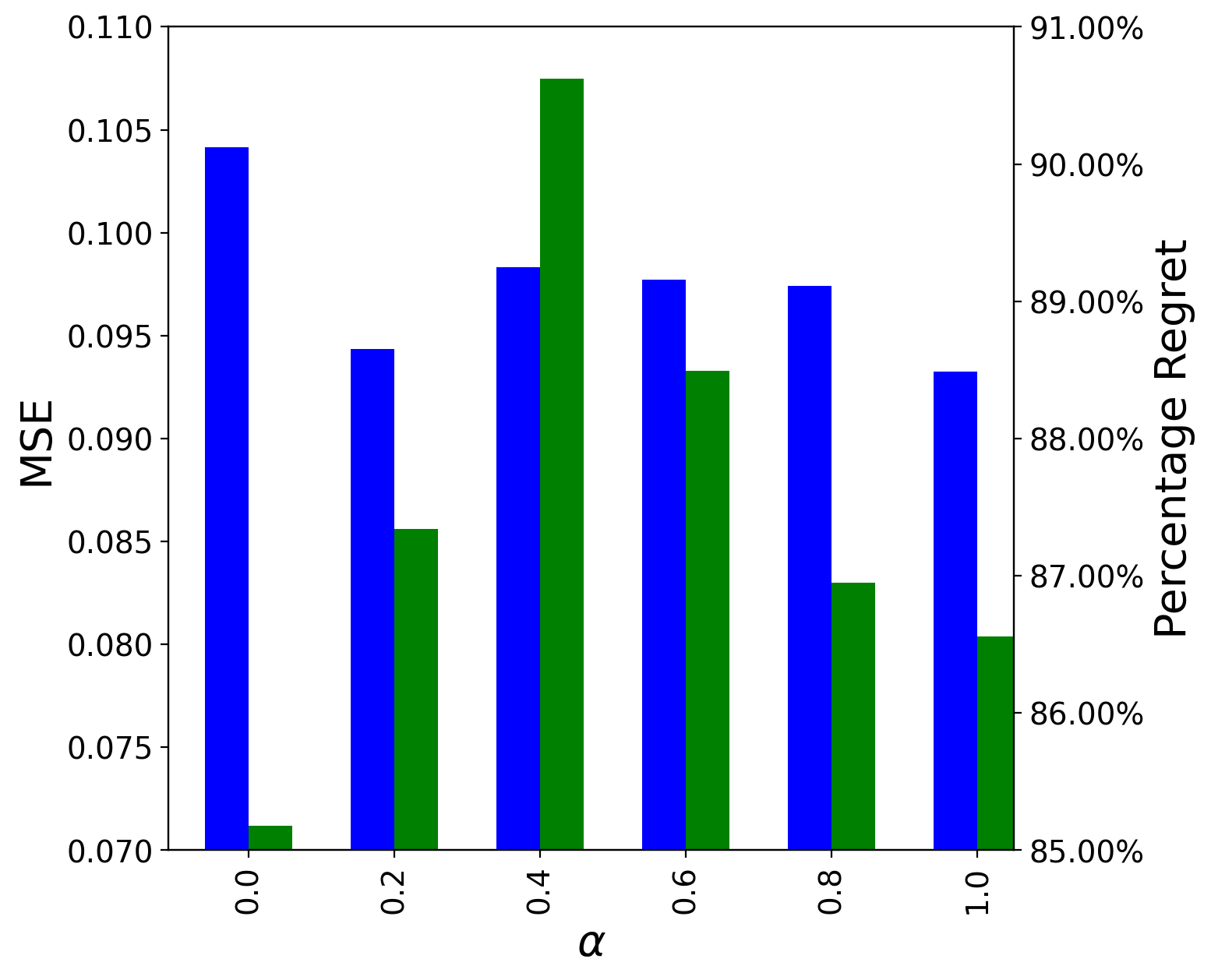}
        \label{}
        \caption{Matching-1}
    \end{subfigure}
    \caption{Change of MSE and Regret with $\alpha$ as a tuning parameter.
    The green and blue bar present the MSE loss and percentage regret respectively.}
    \label{fig:exp_alpha}
\end{figure*}
\paragraph{Diverse Bipartite Matching.} We adopt this from \citet{aaaiFerberWDT20} and \citet{mulamba2020discrete}.
The topologies are taken from the CORA citation network \cite{cora2008}. 
We have 27 disjoint topologies and each of them is considered as an instance. 
The prediction task is to predict which edges are present. Each node has 1433 bag-of-words features. The feature of and edge is formed by concatenating features of the two corresponding nodes. The optimization task is to find the maximum matching after prediction. In the optimization task, diversity constraints ensures that there are some edges between papers of the same field as well as edges between papers of different fields.
We study two instantiations with different degrees of diversity constraints (see Appendix for detail).
\subsection{Experimental Results}
\paragraph{Performance of surrogate loss functions.} In this section, we compare the quality of the rank based loss functions. As described in section~\ref{sect:lossfunctions}, we train the neural network with respect to these loss functions\footnote{The code is available at \url{https://github.com/JayMan91/ltr-predopt}.}. As our objective is to have lower regret, we consider the percentage regret ($\frac{f(v^\star(\hat{c}), c) -  f(v^\star(c), c)}{f(v^\star(c), c)}$) as the evaluation metric.
One way to know whether model training decreases the regret is to monitor the regret on the validation dataset. Figure~\ref{fig:learning_curve} validates that by minimizing the surrogate losses, the neural network models learn to lower the regret. In the shortest path problem, overall, the listwise and pairwise loss functions perform better than the other two. The difference is more pronounced in higher degrees, where the linear model makes larger error. In the scheduling and matching experiments, the pairwise difference loss generalizes the best. The lowest regret of listwise loss is similar to pairwise difference, but its learning curve is less stable.

Next we compare the rank based loss functions with the SPO+ loss~\citep{elmachtoub2021smart}, Blackbox backpropagation  approach~\citep{PogancicPMMR20} and the NCE and MAP approach of \citet{mulamba2020discrete}. We also include a two-stage baseline approach which is trained purely on MSE loss.
We present average regret on test data of 10 independent trials  in Table \ref{tab:Compa}. We refer the readers to \ref{appendix_boxplot} to view the distribution of the regrets.
For the shortest path problem, we again see the regret goes up with the degree parameter. This is expected as the degree parameter controls the magnitude of the errors of the linear model. For degree 4, 6 and 8, the listwise loss function produces the lowest regret. For lower two degrees, the pointwise and pairwise difference loss function produces low regret, but their regret jumps up at higher degrees. This is reasonable as their los functions include the absolute difference between the predicted and true solutions. 
Listwise loss comes second among all in Energy-2 and Matching-1. The pairwise difference loss has lowest regret in Matching-1 and its regret is competitive with the best loss function in these two problems.
In all cases the pointwise loss function performs poorly, even worse than pure MSE (two-stage). The possible explanation is that its added penalty term is not always aligned with regret as discussed in section~\ref{sect:pairwise}. Overall, Table \ref{tab:Compa} and Figure~\ref{fig:learning_curve} indicate the performance of the parirwise difference loss is steady when the underlying predictive model is consistent with the data generation process, whereas the listwise loss is robust and reliable across all problem instances. 

\begin{table*}[ht]
\caption{Impact of $p_{solve}$ on per epoch training time and regret in Energy scheduling instances}
\label{tab:psolve}
\begin{small}
\begin{center}
\label{tab:}
\begin{tabular}{@{}c cc cc cc cc c c@{}}
\toprule
    &   &    \multicolumn{2}{c}{Pointwise}  & \multicolumn{2}{c}{Pairwise}  & \multicolumn{2}{c}{Pairwise-diff}  & \multicolumn{2}{c}{Listwise}             \\ \cmidrule(lr){3-4}\cmidrule(lr){5-6} \cmidrule(lr){7-8} \cmidrule(lr){9-10}
& $p_{solve}$ &   Regret & Time (s) & Regret & Time (s) & Regret & Time (s) & Regret & Time (s)  \\
\toprule
\multirow{ 2}{*}{Energy-1} & 100\% & 2.12\% & 110 & 1.64\% & 110 & 1.63\% & 110 & 1.72\% & 100\\
& 10\% & 2.51\% & 30 & 1.65\% & 30& 1.61\% & 30 & 1.67\% & 25\\
\midrule
\multirow{ 2}{*}{Energy-2} 
& 100\% & 2.70\% & 130 & 2.05\% & 140& 2.07\% &  140 & 1.99\% & 120\\
& 10\% & 2.83\% & 40 & 2.08\% & 45 & 2.04\% & 45 & 1.96\% & 40\\
\bottomrule
\end{tabular}
\end{center}
\end{small}
\end{table*}
\paragraph{MSE vs Regret tradeoff.} 
The three problems we consider happen all to be integer linear programming (ILP) problems. As ILP problems are scale-invariant, the regret can be minimized even if the predictions are multiples of the actual values. 
In Figure \ref{fig:msevsregret}, we show both MSE and regret of the predictions.  In all cases we observe that the MSE loss of the pointwise loss is very low, comparable to minimizing MSE. But its regret is even higher than MSE, as seen before.  The pairwise difference loss function has marginally better MSE than listwise and pairwise losses, showing little added gain from its more MSE-related formulation.

\paragraph{Combined Regression and Ranking Experiment.} The previous experiment reveals that although the performances of the ranking loss functions are comparable with the state of the art in terms of regret, their MSE loss remains high. In this experiment we will consider a convex combination of MSE loss and listwise ranking loss, inspired by ~\cite{10.1145/1835804.1835928} but formulated as a multi-task problem instead of sampling between the two losses:
\begin{equation}
   \alpha  L^l_S(\hat{c},c) + (1-\alpha) \mathbf{mse}(\hat{c},c)
\end{equation}
The motivation behind this loss function is that we can control $\alpha$ as a tuning parameter. Clearly, $\alpha=0$ results in MSE loss whereas $\alpha=1$ results in listwise ranking loss.
Figure~\ref{fig:exp_alpha} displays the impact $\alpha$ on the two scheduling instances. As hypothesised, we see MSE goes up and regret goes down as we increase $\alpha$. This experiment suggests it is possible to choose a value of $\alpha$ so that we sacrifice regret (problem-specific) to attain predictions with lower MSE (close to the true values). 

\paragraph{Impact of $p_{solve}$.} 
The motivation behind ours (and \citet{mulamba2020discrete}) approach is reducing the number of times of the optimization problem is solved during training.
In the next experiment, we show how $p_{solve}$ impacts the model training time. 
Lower values of $p_{solve}$ would make the training faster because of two reasons. Firstly, we have to call the optimization oracle few number of times. Secondly, the smaller cardinality means ranking loss would be computed on fewer datapoints.
In Table \ref{tab:psolve} we report percentage regret and per epoch runtime of the four ranking loss functions for $p_{solve} = 10\% \text{ and } 100 \%$ for the scheduling instances. We choose this problem because this is most difficult optimization problem among the three. Table \ref{tab:psolve} shows that in all cases, reducing $p_{solve}$ to 10\% barely impacts regret for all except pointwise (which already performs worse). On the other hand, the efficiency gain in terms of runtime is significant. 


\section{Conclusion }
We extend the approach of \citet{mulamba2020discrete}, which introduces surrogate loss functions for  \PO \ problems by caching a subset of feasible solutions. Motivated by their work, we frame the  \PO \ problem as a learning to rank problem, where we train models to learn the partial ordering of the pool of feasible solutions. 
This generalizes their approach and inherits its nice properties of separating the solving from the loss computation. We also attempt to draw connections between the two-stage MSE loss and regret minimizing ranking losses.
This stimulates a further cross fertilisation of the rich and more mature field of learning to rank with the relatively recent topic of \PO.
\section*{Acknowledgements}
This research was partially funded by the FWO Flanders project Data-driven logistics (FWO-S007318N), the ANID-Fondecyt Iniciacion grant no 11220864, the European Research Council (ERC) under the European Union’s Horizon 2020 research and innovation programme (Grant No. 101002802, CHAT-Opt) and the Flemish Government under the “Onderzoeksprogramma Artificiele Intelligentie (AI) Vlaanderen” programme.
We are also thankful to the reviewers for their insightful comments. 

\bibliography{mybib}
\bibliographystyle{icml2022}

\newpage
\appendix
\onecolumn

\section{Problem Specification}
\subsection{Shortest Path Problem}
In this problem, we consider shortest path problem on $5 \times 5$ a grid. The goal is to go from the southwest corner to the northeast corner, where we can go towards east or north. We model it as an LP problem.
\begin{align*}
    \min_{x\ge 0} c^\top x \\
    A x = b
\end{align*}
Where $A$ is the incidence matrix.
The decision variable $x$ is binary vector whose entries would be 1 only if corresponding edge is selected for traversal. $b$ is 25 dimensional vector whose first entry is 1 and last  entry is -1 and rest are 0. The constraint must be satisfied to go from the southwest corner to the northeast corner.
With respect to the (predicted) cost vector $c \in \mathbb{R}^{|E|}$, the objective is to minimize the cost. 

\subsection{Energy-cost Aware Scheduling}
In this problem $J$ is the set of tasks to be scheduled on $M$ number of machines maintaining resource requirement of $R$ resources. The tasks must be scheduled over $T$(= 48) set of equal length time periods.
Each task $j$ is specified by its duration $d_j$, earliest start time $e_j$, latest end time $l_j$, power usage $p_j$.$u_{jr}$ is the resource usage of task $j$ for resource $r$ and $c_{mr}$ is the capacity of machine $m$ for resource $r$.
Let $x_{jmt}$ be a binary variable which is 1 only if task $j$ starts at time $t$ on machine $m$.
If $c_t$ is the (predicted) energy price at timeslot $t$, the objective is to minimize the following formulation of total energy cost.
Then the problem can be formulated as  the following MILP:
\begin{align*}
\min_{x_{jmt}} & \sum_{j \in J} \sum_{m \in M} \sum_{t \in T} x_{jmt} \Big( \sum_{t \leq t' < t+d_j} p_j c_{t'} \Big)\\
\text{subject to }& \sum_{m \in M} \sum_{t \in T} x_{jmt} =1 \ , \forall_{j \in J}\\
& x_{jmt} = 0 \ \ \forall_{j \in J} \forall_{m \in M} \forall_{t < e_j}\\
& x_{jmt} = 0 \ \ \forall_{j \in J} \forall_{m \in M} \forall_{t + d_j > l_j} \\
& \sum_{j \in J} \sum_{t - d_{j} < t' \leq t} x_{jmt' } u_{jr}  \leq c_{mr},  \forall_{m \in M} \forall_{r \in R} \forall_{t \in T}
\end{align*}
The first constraint ensures each task is scheduled only once.
The next constraints ensure the task scheduling abides by earliest start time and latest end time constraints.
The final constraint is to respect the resource requirement of the machines.
\subsection{Diverse Bipartite Matching}
We adapt this from \citet{aaaiFerberWDT20}. in this problem 100 nodes representing scientific publications are split into two sets of 50 nodes $N_1$ and $N_2$. A (predicted) reward matrix $c$  indicates the likelihood that a link between each pair of nodes from $N_1$ to $N_2$ exists. Finally a indicator $m_{i,j}$ is $0$ if article $i$ and $j$ share the same subject field, and $0$ otherwise $\forall i \in N_1, j \in N_2$.
Let $p \in [0,1]$ be the rate of pair sharing their field in a solution and $q \in [0,1]$ the rate of unrelated pairs, the goal is to match as much nodes in $N_1$ with at most one node in $N_2$ by selecting edges which maximizes the sum of rewards under similarity and diversity constraints. Formally:
\begin{equation*}
\begin{array}{lll}
\max_x & \sum_{i, j} c_{i, j} x_{i, j} & \\
\text { subject to } & \sum_{j} x_{i, j} \leq 1 & \forall i \in N_{1} \\
& \sum_{i} x_{i, j} \leq 1 & \forall j \in N_{2} \\ 
& \sum_{i . j} m_{i, j} x_{i, j} \geq p \sum_{i, j} x_{i, j} & \\
& \sum_{i . j} (1-m_{i, j}) x_{i, j} \geq q \sum_{i, j} x_{i, j} \\ 
& x_{i,j} \in \{0,1\} & \forall i \in N_1, j \in N_2 \\
\end{array}
\end{equation*}
In our experiments, we considered two variations of this problem with $p=q=$ being $25\%$ and $5\%$, respectively named Matching-1 and Matching-2.

\begin{figure*}[ht]
    \begin{subfigure}[b]{0.33\linewidth}
        \centering
        \includegraphics[scale=0.2]{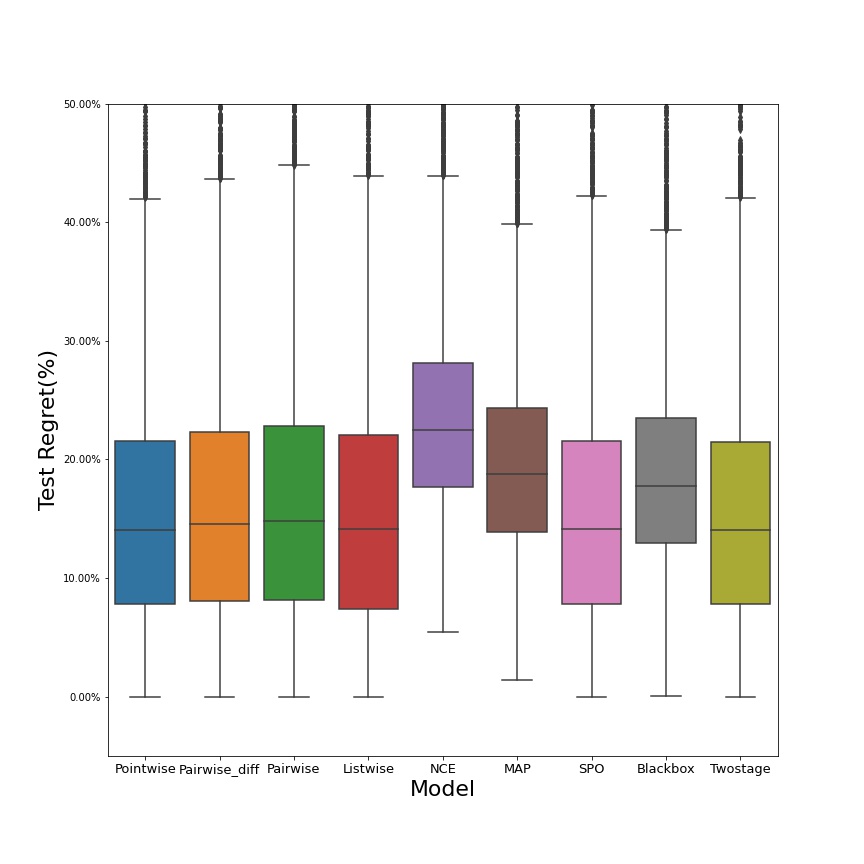}
        \label{}
        \caption{Shortest path (Deg= 1)}
    \end{subfigure}%
    \begin{subfigure}[b]{0.33\linewidth}
        \centering
        \includegraphics[scale=0.2]{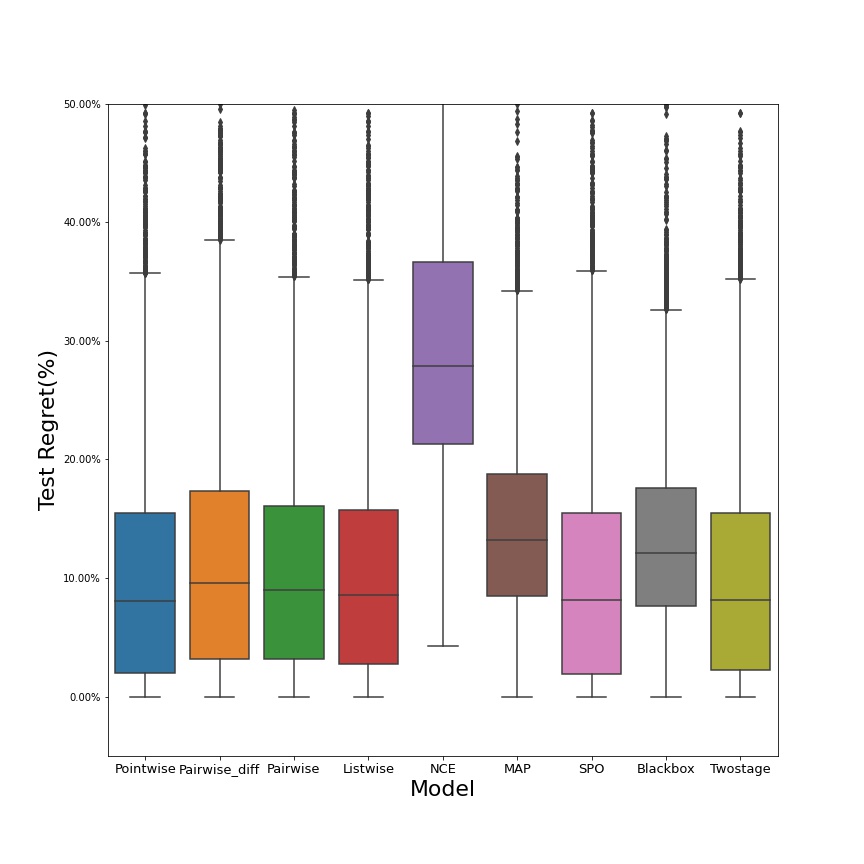}
        \label{}
        \caption{Shortest path (Deg= 2)}
    \end{subfigure}%
    \begin{subfigure}[b]{0.26\linewidth}
        \centering
        \includegraphics[scale=0.2]{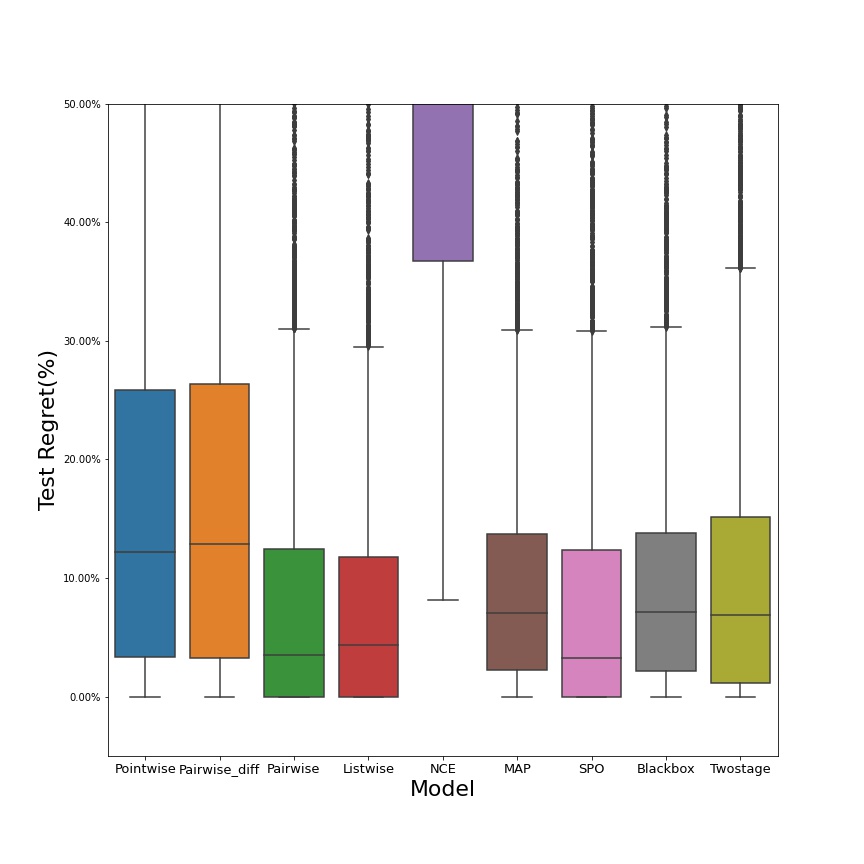}
        \label{}
        \caption{Shortest path (Deg= 4)}
    \end{subfigure}

    \begin{subfigure}[b]{0.33\linewidth}
        \centering
        \includegraphics[scale=0.2]{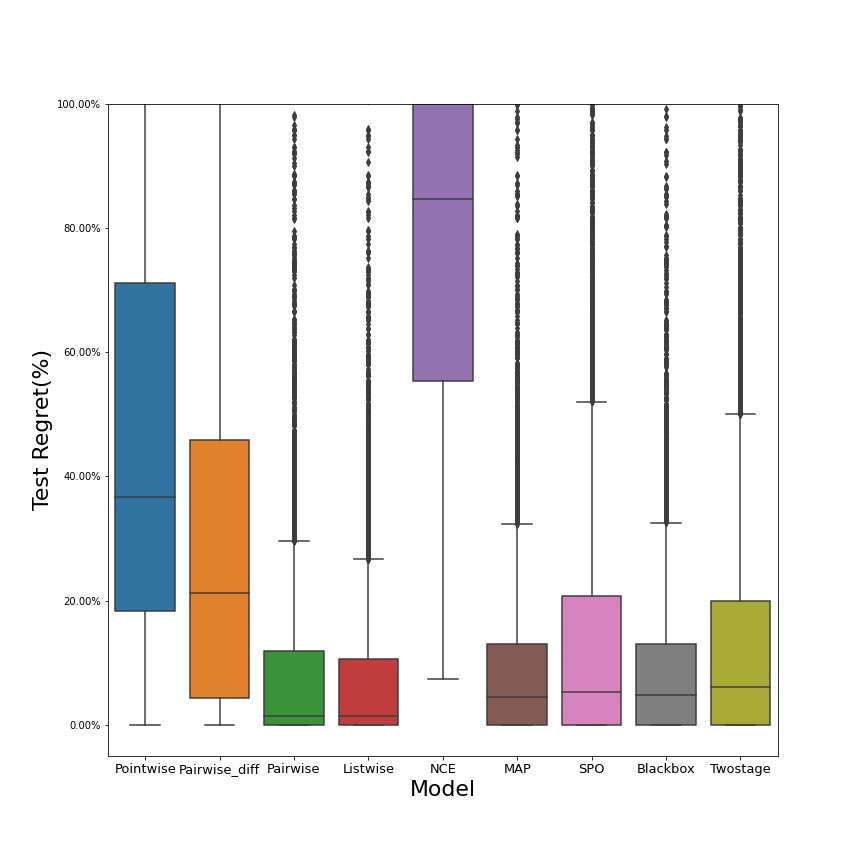}
        \label{}
        \caption{Shortest path (Deg= 6)}
    \end{subfigure}%
    \begin{subfigure}[b]{0.33\linewidth}
        \centering
        \includegraphics[scale=0.2]{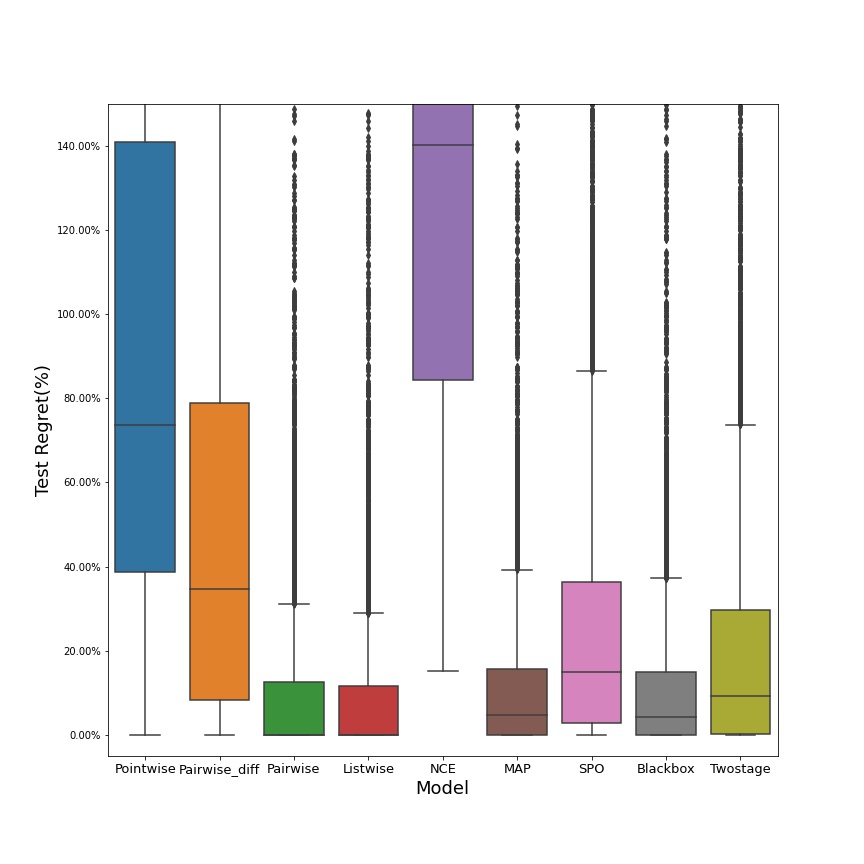}
        \label{}
        \caption{Shortest path (Deg= 8)}
    \end{subfigure}%
    \begin{subfigure}[b]{0.33\linewidth}
        \centering
        \includegraphics[scale=0.2]{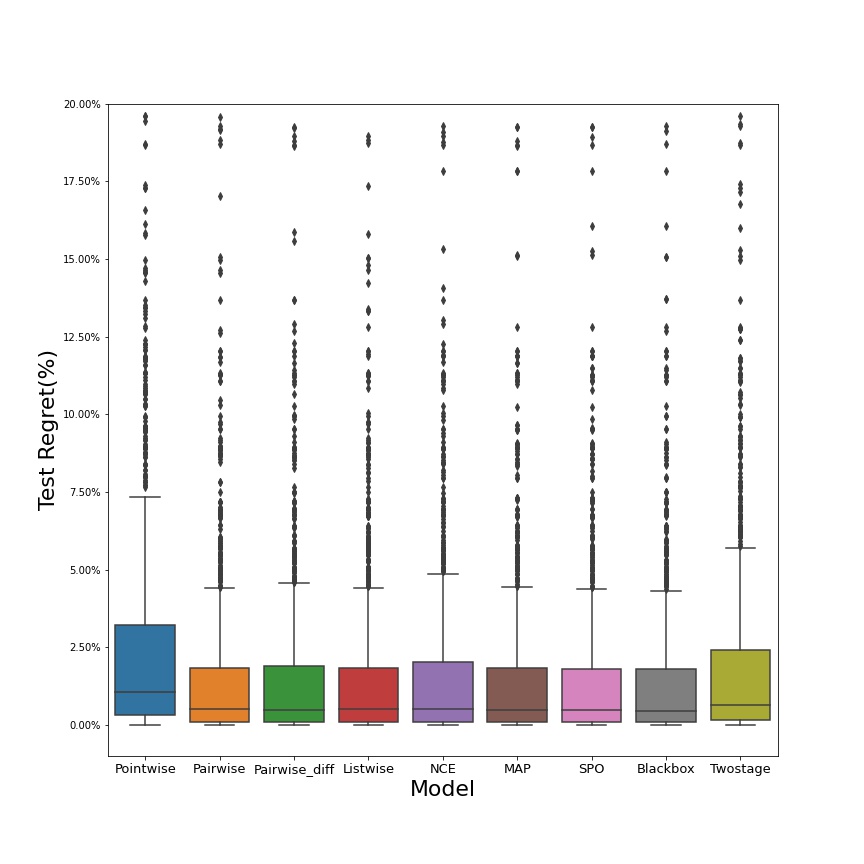}
        \label{}
        \caption{Energy-1}
    \end{subfigure} 

    \begin{subfigure}[b]{0.33\linewidth}
        \centering
        \includegraphics[scale=0.2]{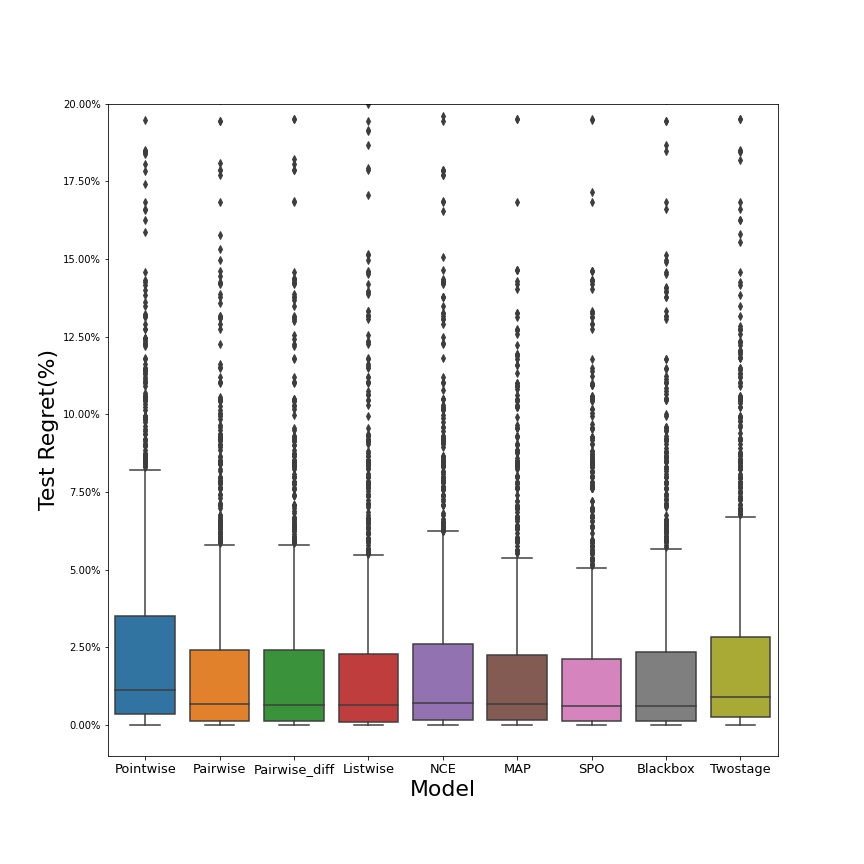}
        \label{}
        \caption{Energy-2}
    \end{subfigure}%
    \begin{subfigure}[b]{0.33\linewidth}
        \centering
        \includegraphics[scale=0.2]{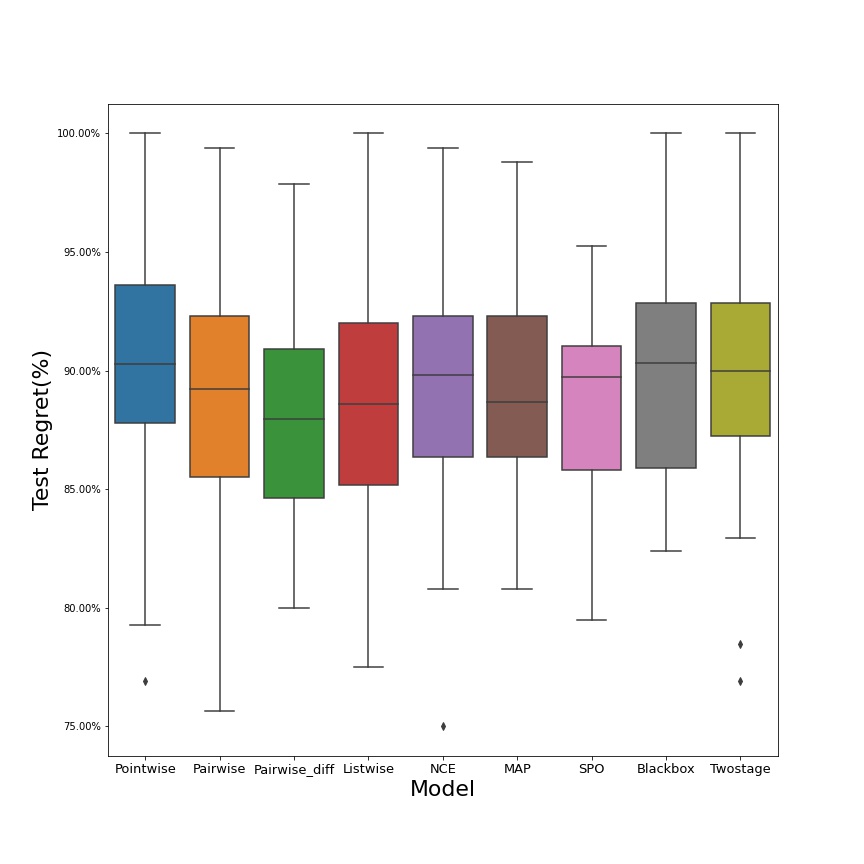}
        \label{}
        \caption{Matching-1}
    \end{subfigure} 
    \begin{subfigure}[b]{0.33\linewidth}
        \centering
        \includegraphics[scale=0.2]{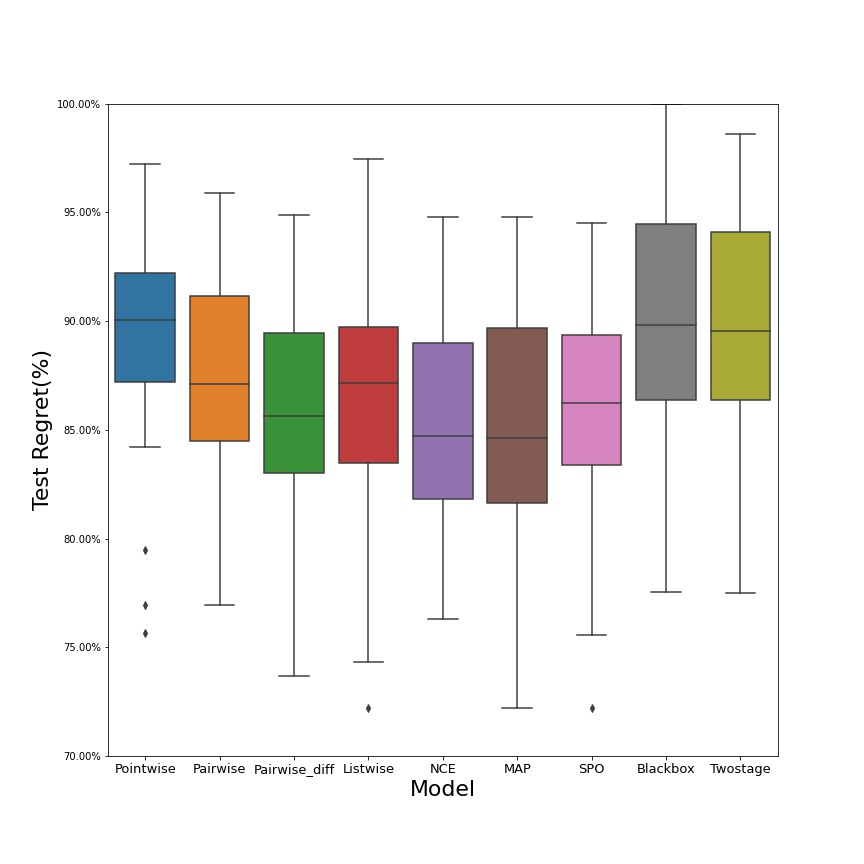}
        \label{}
        \caption{Matching-2}
    \end{subfigure}%
    \caption{Boxplot}
    \label{fig:boxplot}
\end{figure*}
\section{Hyperparameter Configuration}
To reproduce the result reported in this paper use the hyperparameter settings as described in Table~\ref{tab:hyperparams}. 
\begin{table*}[h]
\caption{Details of hyperparameters for reproducing the result}
\label{tab:hyperparams}
\begin{small}
\begin{center}
\label{tab:}
\begin{tabular}{cccc cc cc cc}
\toprule
& & \multicolumn{5}{c}{Shortest Path Problem} \\
\cmidrule(lr){3-7}
    & & Deg-1 &  Deg-2 & Deg-4 & Deg-6 & Deg-8 \\
    \toprule
\multirow{1}{4em}{Pointwise} & learning rate & 0.8 & 0.8 & 0.8 & 0.8 & 0.8 \\
\midrule
\multirow{2}{4em}{Pairwise} & learning rate & 0.1 & 0.1 & 0.1 & 0.1 & 0.1 \\
& margin & 1.0 & 0.1 & 0.1 & 0.1 & 0.1 \\
\midrule
\multirow{2}{4em}{Listwise} & learning rate & 0.1 & 0.1 & 0.1 & 0.1 & 0.1\\
& temperature & 1.0 & 0.1 & 0.05 & 0.05 & 0.05 \\
\midrule
\multirow{1}{10em}{Pairwise-difference} & learning rate & 0.1 & 0.1 & 0.1 & 0.5 & 0.5 \\
\midrule
\multirow{1}{4em}{NCE} & learning rate & 0.5 & 0.5 &0.5 &0.5 &0.5 \\
\midrule
\multirow{1}{4em}{MAP} & learning rate &0.05& 0.05 & 0.05 & 0.7 & 0.7\\
\midrule
\multirow{1}{4em}{SPO} & learning rate & 0.1 & 0.5 & 0.5 & 0.5 & 0.1 \\
\midrule
\multirow{2}{4em}{Blackbox} & learning rate & 0.5 & 0.5 & 0.5 & 0.5 & 0.5 \\
& displace parameter & 1. &1. &1. &1. &1. \\
\midrule
\multirow{1}{4em}{Twostage} & learning rate & 0.5 & 0.5 & 0.1 & 0.5 & 0.5 \\
\midrule
& & \multicolumn{2}{c}{Energy Scheduling} & & \multicolumn{2}{c}{Bipartite Matching} \\
\cmidrule(lr){3-4} \cmidrule(lr){6-7}
    & & Energy-1 & Energy-2  & & Matching-1 & Matching-2\\
    \toprule
\multirow{1}{4em}{Pointwise} & learning rate &  0.9 & 0.5 & & 0.01 & 0.01\\
\midrule
\multirow{2}{4em}{Pairwise} & learning rate &  0.1 & 0.1 & & 0.001 & 0.01\\
& margin & 100 & 100 & &0.2 & 1.0\\
\midrule
\multirow{2}{4em}{Listwise} & learning rate & 0.1 & 0.1 && 0.005 & 0.01\\
& temperature & 1.0 & 1.0 &  & 0.01 & 1.0\\
\midrule
\multirow{1}{10em}{Pairwise-difference} & learning rate &  0.5 & 0.5 & & 0.01 & 0.01\\
\midrule
\multirow{1}{4em}{NCE} & learning rate & 0.5 & 0.5 &&  0.001 & 0.01\\
\midrule
\multirow{1}{4em}{MAP} & learning rate & 0.5 & 0.5 && 0.001 & 0.01\\
\midrule
\multirow{1}{4em}{SPO} & learning rate &  0.9 & 0.9 && 0.01 & 0.01\\
\midrule
\multirow{2}{4em}{Blackbox} & learning rate & 0.1 & 0.1 && 0.01 & 0.01\\
& displace parameter &  0.001 & 0.1 && $10^{-5}$ & 0.001\\
\midrule
\multirow{1}{4em}{Twostage} & learning rate & 0.7 & 0.7 && 0.01 & 0.01 \\
\bottomrule
\end{tabular}
\end{center}
\end{small}
\end{table*}
\section{Distribution of the regret of the predictions}
In Table \ref{tab:Compa}, we report only the average regret of all the instances over 10 independent trails. 
In Figure {fig:boxplot}, we show the distributions of the regret across all the test instances with box-plots  for the three problems. 
\label{appendix_boxplot}
\end{document}